\documentclass[10pt,twocolumn,letterpaper]{article}

\usepackage{cvpr}              

\definecolor{cvprblue}{rgb}{0.21,0.49,0.74}
\usepackage{xcolor}         
\usepackage{amsmath}
\usepackage{bm}
\usepackage{color, colortbl}
\usepackage{multirow}  
\definecolor{front-color}{HTML}{F5FFFA}
\definecolor{Gray}{gray}{0.90}
\usepackage[pagebackref,breaklinks,colorlinks,allcolors=cvprblue]{hyperref}

\def\paperID{13563} 
\def\confName{CVPR}
\def\confYear{2025}

\title{GroupMamba: Efficient Group-Based Visual State Space Model}

\author{%
    Abdelrahman Shaker$^{1}$ \quad
    Syed Talal Wasim$^{2,3}$ \quad
    Salman Khan$^{1}$ \\
    Juergen Gall$^{2,3}$ \quad
    Fahad Shahbaz Khan$^{1,4}$ \\[5pt]
    $^{1}$Mohamed Bin Zayed University of Artificial Intelligence \quad
    $^{2}$University of Bonn \\
    $^{3}$Lamarr Institute for Machine Learning and Artificial Intelligence \quad
    $^{4}$Linköping University
}

\begin{document}
\maketitle
\begin{abstract}
State-space models (SSMs) have recently shown promise in capturing long-range dependencies with subquadratic computational complexity, making them attractive for various applications. However, purely SSM-based models face critical challenges related to stability and achieving state-of-the-art performance in computer vision tasks. Our paper addresses the challenges of scaling SSM-based models for computer vision, particularly the instability and inefficiency of large model sizes. We introduce a parameter-efficient modulated group mamba layer that divides the input channels into four groups and applies our proposed SSM-based efficient Visual Single Selective Scanning (VSSS) block independently to each group, with each VSSS block scanning in one of the four spatial directions. The Modulated Group Mamba layer also wraps the four VSSS blocks into a channel modulation operator to improve cross-channel communication.
Furthermore, we introduce a distillation-based training objective to stabilize the training of large models, leading to consistent performance gains. Our comprehensive experiments demonstrate the merits of the proposed contributions, leading to superior performance over existing methods for image classification on ImageNet-1K, object detection, instance segmentation on MS-COCO, and semantic segmentation on ADE20K. 
Our tiny variant with 23M parameters achieves state-of-the-art performance with a classification top-1 accuracy of 83.3\% on ImageNet-1K, while being 26\% efficient in terms of parameters, compared to the best existing Mamba design of same model size. Code and models are available at: \hyperlink{https://github.com/Amshaker/GroupMamba}{https://github.com/Amshaker/GroupMamba}

\end{abstract}    
\section{Introduction}
\label{sec:intro}
\begin{figure}
  \centering
    \includegraphics[width=0.45\textwidth]
    {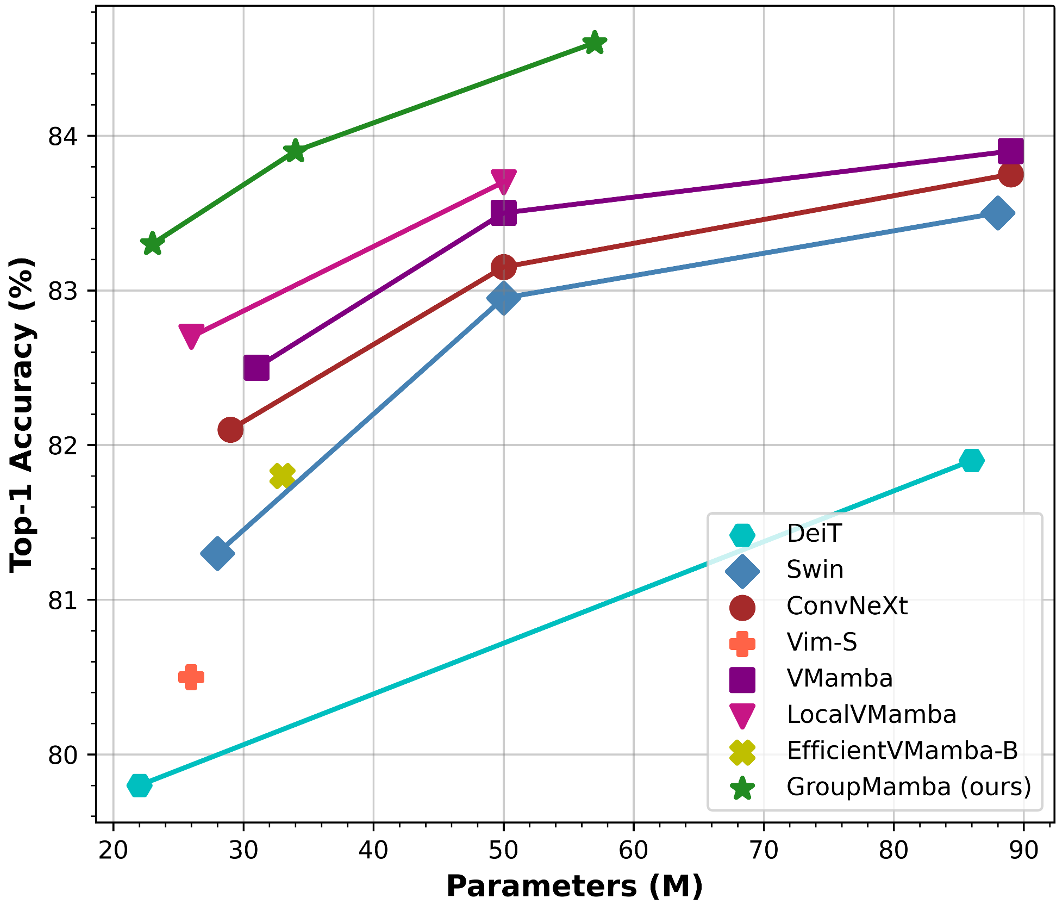}

    \caption{Comparison in terms of Parameters vs. Top-1 Accuracy on ImageNet-1k~\citep{deng2009imagenet}. Our GroupMamba-B achieves superior top-1 classification accuracy while reducing parameters by 36\% compared to VMamba~\citep{yue2024vmamba}.}
   
    \label{fig:intro_figure}
\end{figure}
Various context modeling methods have emerged in the domains of language and vision understanding. These include Convolution~\cite{he2016resnet, yang2022focalnets}, Attention~\cite{vaswani2017attention}, and, more recently, State Space Models~\cite{gu2022s4, gu2023mamba}. Transformers with their multi-headed self-attention mechanism~\cite{vaswani2017attention} have been central to both language models such as GPT-3~\cite{brown2020gpt3} and vision models such as Vision Transformers~\cite{dosovitskiy2021vit, liu2021Swin}. However, challenges arose due to the quadratic computational complexity of attention mechanisms particularly for longer sequences, leading to the recent emergence of State Space models such as S4~\cite{gu2022s4}.

While being effective in handling extended input sequences due to their linear complexity in terms of sequence lengths, S4~\cite{gu2022s4} encountered limitations in global context processing in information-dense data, especially in domains like computer vision due to the data-independent nature of the model. Alternatively, approaches such as global convolutions-based state space models~\cite{fu2023flashfftconv} and Liquid S4~\cite{hasani2022liquid} have been proposed to mitigate the aforementioned limitations. The recent Mamba~\cite{gu2023mamba} introduces the S6 architecture which aims to enhance the ability of state-space models to handle long-range dependencies efficiently. The selective-scan algorithm introduced by Mamba uses input-dependent state-space parameters, which allow for better in-context learning while still being computationally efficient compared to self-attention.

However, Mamba, specifically the S6 algorithm, is known to be unstable for e.g., image classification, especially when scaled to large sizes~\cite{patro2024simba}. Additionally, the Mamba model variant used in image classification, generally called the VSS (Visual State Space) block, can be more efficient in terms of parameters and compute requirements based on the number of channels. The VSS block includes extensive input and output projections along with depth-wise convolutions, whose parameters and compute complexities are directly proportional to the number of channels in the input. To address this issue, we propose a hierarchical-based \emph{Modulated Group Mamba} layer that mitigates the aforementioned issues in a computation and parameter-efficient manner. The main contributions of our paper are:

\begin{enumerate}
    \item We introduce a \emph{Modulated Group Mamba} layer, inspired by Group Convolutions, which enhances computational efficiency and interaction in state-space models by using a multi-direction scanning method for comprehensive spatial coverage and effective modeling of local and global information.
    \item We introduce a \emph{Channel Affinity Modulation (CAM)} operator, which enhances communication across channels to improve feature aggregation, addressing the limited interaction inherent in the grouping operation.
    \item To address the instability issue in the SSM-based architecture, we introduce a distillation-based training objective designed to stabilize models with a large number of parameters, leading to better performance and a smooth loss convergence trend.
    \item We build a series of parameter-efficient generic classification models called “GroupMamba”, based on the proposed \emph{Modulated Group Mamba} layer. Our \textit{tiny} variant achieves 83.3\% top-1 accuracy on ImageNet-1k~\cite{deng2009imagenet} with $23M$ parameters and $4.5G$ FLOPs. Additionally, our \textit{base} variant achieves top-1 accuracy of 84.5\% with $57M$ parameters and $14G$ FLOPs, outperforming all recent SSM methods (see Fig.~\ref{fig:intro_figure}).

\end{enumerate}

\section{Related Work}
\label{sec:related}

Convolutional Neural Networks (ConvNets) have been the popular choice for computer vision tasks since the introduction of AlexNet~\citep{alex2012alexnet}. The field has rapidly evolved with several landmark ConvNet architectures~\citep{simonyan2014vgg,szegedy2015inception, he2016resnet, howard2017mobilenet, tan2019efficientnet}. Alongside these architectural advances, significant efforts have been made to refine individual convolution layers, including depthwise convolution~\citep{depthwiseconv}, group convolution~\citep{group_conv}, and deformable convolution~\citep{deformable_conv}. Recently, ConvNeXt variants~\citep{liu2022convnext, convnext_v2} have taken concrete steps towards modernizing traditional 2D ConvNets by incorporating macro designs with advanced settings and training recipes to achieve on-par performance with the state-of-the-art models.

In recent years, the pioneering Vision Transformer (ViT)~\citep{dosovitskiy2021vit} has significantly impacted the computer vision field, including tasks such as image classification~\citep{touvron2021deit, liu2021Swin, liu2022swinv2, MViT}, object detection~\citep{DETR, zhu2020deformable, meng2021conditional, zhang2022dino}, and segmentation~\citep{MaskFormer, MAVOS, SAM}. ViT~\citep{dosovitskiy2021vit} introduces a monolithic design that approaches an image as a series of flattened 2D patches without image-specific inductive bias. The remarkable performance of ViT for computer vision tasks, along with its scalability, has inspired numerous subsequent endeavors to design better architectures. The early ViT-based models usually require large-scale datasets (e.g., JFT-300M~\citep{JFT}) for pretraining. Later, DeiT~\citep{touvron2021deit} proposes advanced training techniques in addition to integrating a distillation token into the architecture, enabling effective training on smaller datasets (e.g., ImageNet-1K~\citep{deng2009imagenet}). Since then, subsequent studies have designed hierarchical and hybrid architectures by combining CNN and ViT modules to improve performance on different vision tasks \citep{BoTNet, EdgeNeXt,convit,swiftformer, MViT}. Another line of work is to mitigate the quadratic complexity inherent in self-attention, a primary bottleneck of ViTs. This effort has led to significant improvements and more efficient and approximated variants~\citep{LinFormer, swiftformer, EdgeViT, MobileViT2, ReFormer, Twins, MaxViT}, offering reduced complexity while maintaining effectiveness.

Recently, State Space Models (SSMs) have emerged as an alternative to ViTs~\citep{vaswani2017attention}, capturing the intricate dynamics and inter-dependencies within language sequences~\citep{gu2022s4}. One notable method in this area is the structured state-space sequence model (S4)~\citep{gu2022s4}, designed to tackle long-range dependencies while maintaining linear complexity. Following this direction, several models have been proposed, including S5~\citep{smith2023s5}, H3~\citep{fu2023h3}, and GSS~\citep{mehta2022gss}. More recently, Mamba~\citep{gu2023mamba} introduces an input-dependent SSM layer and leverages a parallel selective scan mechanism (S6).

In the visual domain, various works have applied SSMs to different tasks. In particular for image classification, VMamba~\citep{yue2024vmamba} uses Mamba with bidirectional scans across both spatial dimensions in a hierarchical Swin-Transformer~\citep{liu2021Swin} style design to build a global receptive field efficiently. A concurrent work, Vision Mamba (Vim)~\citep{lianghui2024vim}, instead proposed a monolithic design with a single bidirectional scan for the entire image, outperforming traditional vision transformers like DeiT. LocalVMamba~\citep{huang2024localmamba} addresses the challenge of capturing detailed local information by introducing a scanning methodology within distinct windows (inspired from Swin-Transformer~\citep{liu2021Swin}), coupled with dynamic scanning directions across network layers. EfficientVMamba~\citep{pei2024efficientvmamba} integrates atrous-based selective scanning and dual-pathway modules for efficient global and local feature extraction, achieving competitive results with reduced computational complexity. 
These models have been applied for image classification, as well as image segmentation~\citep{liu2024swin_umamba, ma2024umamba, ruan2024vmunet, gong2024nnmamba}, video understanding~\citep{yang2024vivim, li2024videomamba, chen2024video}, and various other tasks~\citep{guo2024mambamorph, he2024pan, wang2024graph, guo2024mambair, liang2024pointmamba}. Their wide applicability shows the effectiveness of SSMs~\citep{gu2022s4, smith2023s5, fu2023h3, mehta2022gss}, and in particular Mamba~\citep{gu2023mamba}, in the visual domain. In this paper, we propose a \emph{Modulated Group Mamba} layer that mitigates the drawbacks of the default vision Mamba block, such as lack of stability~\citep{patro2024simba} and the increased number of parameters with respect to the number of channels.
\section{Method}
\label{sec:method}

\begin{figure*}
  \centering
    \includegraphics[width=0.95\textwidth]
    {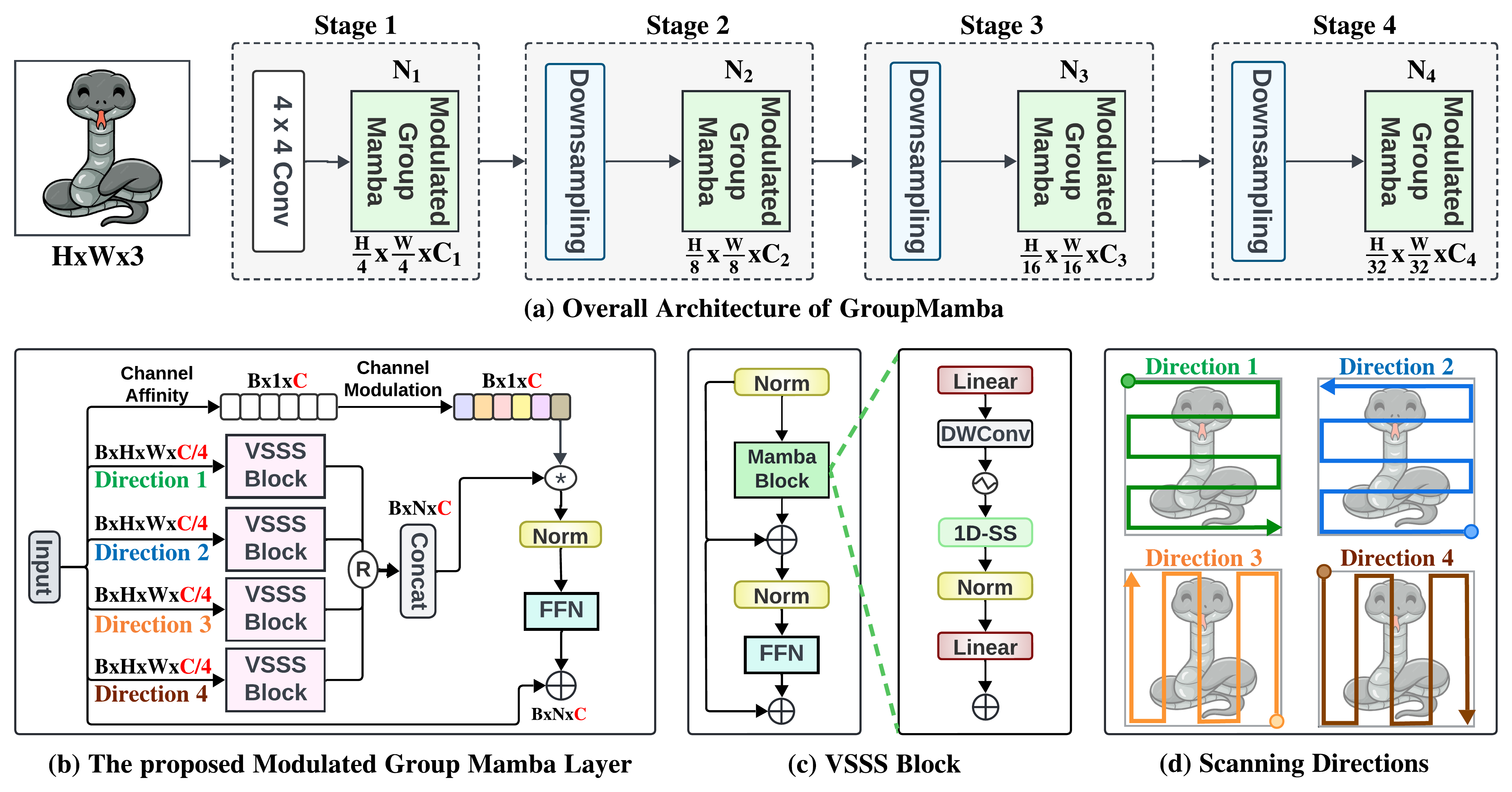}

    \caption{Overview of the proposed method. \textbf{Top Row:} The overall architecture of our framework with a consistent hierarchical design comprising four stages. 
    \textbf{Bottom Row:} We present \textbf{(b)} The design of the modulated group mamba layer. The input channels are divided into four groups with a single scanning direction for each VSSS block. This significantly reduces the computational complexity compared to the standard mamba layer, with similar performance. Channel Affinity Modulation mechanism is introduced to address the limited interactions within the VSSS blocks. \textbf{(c)} The design of VSSS block. It consists of Mamba block with 1D Selective Scanning block followed by FFN. \textbf{(d)} The four scanning directions used for the four VSSS blocks are illustrated.}

    \label{fig:method_overall}
\end{figure*}

\textbf{Motivation:} Our method is motivated based on the observations with respect to the limitations of existing Visual State-Space models. 

\begin{itemize}
    \item \textit{Lack of Stability for Larger Models}: We observe from~\cite{patro2024simba} that Mamba~\citep{gu2023mamba} based image classification models with an MLP channel mixer are unstable when scaled to a large number of parameters. This instability can be seen in SiMBA-L (MLP)~\citep{patro2024simba}, which leads to sub-optimal classification results of $49\%$ accuracy. We mitigate this issue by introducing a \emph{Modulated Group Mamba} design alongside a distillation objective (as presented in Sec.~\ref{sec:method:loss}) that stabilizes the Mamba SSM training without modifying the channel mixer.
    \item \textit{Efficient Improved Interaction}: Given the computational impact of Mamba-based design on the number of channels, the proposed \emph{Modulated Group Mamba} layer is computationally inexpensive and parameter efficient than the default Mamba and able to model both local and global information from the input tokens through multi-direction scanning. An additional \emph{Channel Affinity Modulation} operator is proposed in this work to compensate for the limited channel interaction due to the grouped operation and enhance their interactions.
\end{itemize}

\subsection{Preliminaries}

\textbf{State-Space Models:} State-space models (SSMs) like S4~\citep{gu2022s4} and Mamba~\citep{gu2023mamba} are structured sequence architectures inspired by a combination of recurrent neural networks (RNNs) and convolutional neural networks (CNNs), with linear or near-linear scaling in sequence length. Derived from continuous systems, SSMs define and 1D \textit{function-to-function map} for an input $x(t) \in \mathbb{R}^L \rightarrow y(t) \in \mathbb{R}^L $ via a hidden state $h(t) \in \mathbb{R}^N$. More formally, SSMs are described by the continuous time Ordinary Differential Equation (ODE) in Eq.~\ref{eq:meth:ssm}.
\begin{equation}
\begin{split}
    h'(t) &= {\mathbf A}h(t) + {\mathbf B}x(t), \\
    y(t) &= {\mathbf C}h(t),
    \label{eq:meth:ssm}
\end{split}
\end{equation}
where $h(t)$ is the current hidden state, $h'(t)$ is the updated hidden state, $x(t)$ is the current input, $y(t)$ is the output, ${\mathbf A} \in \mathbb{R}^{N\times N}$ is SSM's evolution matrix, and ${\mathbf B} \in \mathbb{R}^{1\times N}, {\mathbf C} \in \mathbb{R}^{ 1\times N}$ are the input and output projection matrices, respectively.

\noindent\textbf{Discrete State-Space Models:} To allow these models to be used in sequence modeling tasks in deep learning, they need to be discretized, converting the SSM from a continuous time \textit{function-to-function map} into a discrete-time \textit{sequence-to-sequence map}. S4~\citep{gu2022s4} and Mamba~\citep{gu2023mamba} are among the discrete adaptations of the continuous system, incorporating a timescale parameter ${\mathbf \Delta}$ to convert the continuous parameters ${\mathbf A}, {\mathbf B}$ into their discrete equivalents $\overline{{\mathbf A}}, \overline{{\mathbf B}}$. This discretization is typically done through the Zero-Order Hold (ZOH) method given in Eq.~\ref{eq:meth:discrete_ssm}.
\begin{equation}
\begin{split}
\overline{{\mathbf A}} &= \exp({\mathbf \Delta \mathbf A}), \\
\overline{{\mathbf B}} &= ({\mathbf \Delta \mathbf A})^{-1} (\exp({\mathbf \Delta \mathbf A}) - {\mathbf I}) \cdot {\mathbf \Delta \mathbf B} \\
h_t &= \overline{{\mathbf A}} h_{t-1} + \overline{{\mathbf B}} x_t, \\
y_t &= {\mathbf C}h_t.
\label{eq:meth:discrete_ssm}
\end{split}
\end{equation}

While both S4~\citep{gu2022s4} and Mamba~\citep{gu2023mamba} utilize a similar discretization step as stated above in Eq.~\ref{eq:meth:discrete_ssm}, Mamba differentiates itself from S4 by conditioning the parameters ${\mathbf \Delta} \in \mathbb{R}^{B \times L \times D}$, ${\mathbf B} \in \mathbb{R}^{B\times L \times N}$ and ${\mathbf C} \in \mathbb{R}^{B\times L \times N}$, on the input $x \in \mathbb{R}^{B \times L \times D}$, through the S6 Selective Scan Mechanism, where $B$ is the batch size, $L$ is the sequence length, and $D$ is the feature dimension.

\subsection{Overall Architecture}

As shown in Fig.~\ref{fig:method_overall} (a), our model uses a hierarchical architecture, similar to Swin-Transformer~\citep{liu2021Swin}, with four stages to efficiently process images at varying resolutions. Assuming an input image, $\mathbf{I} \in \mathbb{R}^{H \times W \times 3}$, we first apply a $\textsf{Patch Embedding}$ layer to divide the image into non-overlapping patches of size $4 \times 4$ and embed each patch into a $C_1$-dimensional feature vector. The patch embedding layer is implemented using two $3 \times 3$ convolutions with a stride of 2. This produces features maps of size $\frac{H}{4} \times \frac{W}{4} \times C_{1}$ at the first stage. These feature maps are passed through $N_{1}$ blocks of our Modulated Grouped Mamba (as detailed in Sec.~\ref{sec:method:MGM}). In each subsequent stage, a down-sampling layer merges patches in a $2 \times 2$ region, followed by another $N$ blocks of our Modulated Grouped Mamba layer. Hence, feature size at stages two, three and four are $\frac{H}{8} \times \frac{W}{8} \times C_{2}$, $\frac{H}{16} \times \frac{W}{16} \times C_{3}$, and $\frac{H}{32} \times \frac{W}{32} \times C_{4}$, respectively. 

\subsection{Modulated Group Mamba Layer}
\label{sec:method:MGM}

We present the overall operations of the proposed \emph{Modulated Group Mamba} layer (Fig.~\ref{fig:method_overall} (b)) for an input sequence $\mathbf{X}_{\textsf{in}}$, with dimensions $(B, H, W, C)$, where $B$ is the batch size, $C$ is the number of input channels and $H$/$W$ are the width and height of the feature map, in Eq.~\ref{eq:complete}. 
\begin{equation}
\begin{aligned}
    \mathbf{X}_{\textsf{GM}} &= \textsf{GroupedMamba}(\mathbf{X}_{\textsf{in}}, \Theta) \\
    \mathbf{X}_{\textsf{CAM}} &= \textsf{CAM}(\mathbf{X}_{\textsf{GM}}, \textsf{Affinity}(\mathbf{X}_{\textsf{in}})) \\
    \mathbf{X}_{\textsf{out}} &= \mathbf{X}_{\textsf{in}} + \textsf{FFN}(\textsf{LN}(\mathbf{X}_{\textsf{CAM}}))
    \label{eq:complete}
\end{aligned}
\end{equation}
Here, $\mathbf{X}_{\textsf{GM}}$ is the output of Eq.~\ref{eq:groupedmamba}, $\mathbf{X}_{\textsf{CAM}}$ is the output of Eq.~\ref{eq:recalib}, $\textsf{LN}$ is the Layer Normalization~\citep{ba2016layer} operation, $\textsf{FFN}$ is the Feed-Forward Network as described by Eq.~\ref{eq:ffn}, and $\mathbf{X}_{\textsf{out}}$ is the final output of the Modulated Group Mamba block. The individual operations, namely the $\textsf{GroupedMamba}$ operator, the $\textsf{VSSS}$ block used inside the $\textsf{GroupedMamba}$ operator, and the $\textsf{CAM}$ operator, are presented in Sec.~\ref{sec:method:MGM:VSSS}, Sec.~\ref{sec:method:MGM:GM} and Sec.~\ref{sec:method:MGM:CAM}, respectively.

\subsubsection{Visual Single Selective Scan (VSSS) Block}
\label{sec:method:MGM:VSSS}

The VSSS block (Fig.~\ref{fig:method_overall} (c)) is a token and channel mixer based on the Mamba operator, comprising of a Mamba block followed by a Feed-Forward Network, each with a LayerNorm before it. Mathematically, for an input token sequence $\mathbf{Z}_{\textsf{in}}$, the VSSS block performs the operations as described in Eq.~\ref{eq:vsss}.
\begin{equation}
\begin{aligned}
    \mathbf{Z}'_{\textsf{out}} &= \mathbf{Z}_{\textsf{in}} + \textsf{Mamba}(\textsf{LN}(\mathbf{Z}_{\textsf{in}})) \\
    \mathbf{Z}_{\textsf{out}} &= \mathbf{Z}'_{\textsf{out}} + \textsf{FFN}(\textsf{LN}(\mathbf{Z}'_{\textsf{out}}))
    \label{eq:vsss}
\end{aligned}
\end{equation}
Where $\mathbf{Z}_{\textsf{out}}$ is the output sequence, $\textsf{Mamba}$ is the discretized Mamba SSM operator as described in Eq.~\ref{eq:meth:discrete_ssm}.
\begin{equation}
    \textsf{FFN}(\textsf{LN}(\mathbf{Z}'_{\textsf{out}})) = \textsf{GELU}(\textsf{LN}(\mathbf{Z}'_{\textsf{out}}) \mathbf{W}_1 + \mathbf{b}_1) \mathbf{W}_2 + \mathbf{b}_2
    \label{eq:ffn}
\end{equation}
Where $\textsf{GELU}$~\citep{hendrycks2023gelu} is the activation function and $\mathbf{W}_1$, $\mathbf{W}_2$, $\mathbf{b}_1$, and $\mathbf{b}_2$ are weights and biases for the linear projections.

\subsubsection{Grouped Mamba Operator}
\label{sec:method:MGM:GM}

Considering the motivation presented earlier in Sec.~\ref{sec:method}, we aim to design a variant of the Mamba~\citep{gu2023mamba} that is both computationally efficient and can effectively model the spatial dependencies of the input sequence. Given that Mamba is computationally inefficient on large number of channels $C$ in the input sequence, we propose a grouped variant of the operator, inspired by Grouped Convolutions. The Grouped Mamba operation is a variant of the VSSS block presented in Sec.~\ref{sec:method:MGM:VSSS}, where the input channels are divided into groups, and the VSSS operator is applied separately to each group. Specifically, we divide the input channels into four groups, each of size $\frac{C}{4}$, and an independent VSSS block is applied to each group. Hence, the proposed grouped mamba operator enhances the model efficiency by splitting channels into smaller groups. To better model spatial dependencies in the input, each of the four groups scans in one of four directions across the input: left-to-right, right-to-left, bottom-to-top, and top-to-bottom as outlined in Fig.~\ref{fig:method_overall} (d).

Let $G=4$ be the number of groups representing four scanning directions: left-to-right, right-to-left, top-to-bottom, and bottom-to-top. We form four sequences from the input sequence $\mathbf{X}_{\textsf{in}}$, namely $\mathbf{X}_{\textsf{LR}}$, $\mathbf{X}_{\textsf{RL}}$, $\mathbf{X}_{\textsf{TB}}$, and $\mathbf{X}_{\textsf{BT}}$, each of shape $(B, H, W, \frac{C}{4})$, representing one of the four directions specified earlier. These are then flattened to form a single token sequence of shape $(B, N, \frac{C}{4})$, where $N=W\times H$ is the number of tokens in the sequence. The parameters for each of the four groups can be specified by $\theta_{\textsf{LR}}$, $\theta_{\textsf{RL}}$, $\theta_{\textsf{TB}}$, and $\theta_{\textsf{BT}}$, respectively, for each of the four groups, representing the parameters for the VSSS blocks.

Given the above definitions, the overall relation for the Grouped Mamba operator can be written as shown in Eq.~\ref{eq:groupedmamba}.
\begin{equation}
\begin{aligned}
    \mathbf{X}_{\textsf{GM}} = 
    &\textsf{GroupedMamba}(\mathbf{X}_{\textsf{in}}, \Theta) = 
    \textsf{Concat}\big( \\
    & \textsf{VSSS}(\mathbf{X}_{\textsf{LR}}, \Theta_{\textsf{LR}}),\,
    \textsf{VSSS}(\mathbf{X}_{\textsf{RL}}, \Theta_{\textsf{RL}}), \\
    & \textsf{VSSS}(\mathbf{X}_{\textsf{TB}}, \Theta_{\textsf{TB}}),\,
    \textsf{VSSS}(\mathbf{X}_{\textsf{BT}}, \Theta_{\textsf{BT}}) \big)
    \label{eq:groupedmamba}
\end{aligned}
\end{equation}
Where:
\begin{itemize}
  \item $\mathbf{X}_{\textsf{LR}}$, $\mathbf{X}_{\textsf{RL}}$, $\mathbf{X}_{\textsf{TB}}$, and $\mathbf{X}_{\textsf{BT}}$ represent the input tensors scanned in the respective directions.
  \item $\Theta_{\textsf{LR}}$, $\Theta_{\textsf{RL}}$, $\Theta_{\textsf{TB}}$, and $\Theta_{\textsf{BT}}$ represents the parameters of the VSSS block for each direction.
  \item The output of each Mamba operator is reshaped again to $(B, H, W, \frac{C}{4})$, and concatenated back to form the token sequence $\mathbf{X}_{\textsf{GM}}$, again of the size $(B, H, W, C)$.
\end{itemize}

\subsubsection{Channel Affinity Modulation (CAM)}
\label{sec:method:MGM:CAM}

On its own, the Grouped Mamba operator may have a disadvantage in the form of limited information exchange across channels, given the fact that each operator in the group only operates over $\frac{C}{4}$ channels. To encourage the exchange of information across channels, we propose a \empty{Channel Affinity Modulation} operator, which recalibrates channel-wise feature responses to enhance the representation power of the network. In this block, we first average pool the input to calculate the channel statistics as shown in Eq.~\ref{eq:channelstat}.
\begin{equation}
    \textsf{ChannelStat}(\mathbf{X}_{\textsf{in}}) = \textsf{AvgPool}(\mathbf{X}_{\textsf{in}})
    \label{eq:channelstat}
\end{equation}
where $\mathbf{X}_{\textsf{in}}$ is the input tensor, and $\textsf{AvgPool}$ represents the global average pooling operation. Next comes the affinity calculation operation as shown in Eq.~\ref{eq:affinity}.
\begin{equation}
    \textsf{Affinity}(\mathbf{X}_{\textsf{in}}) = \sigma\left(W_2 \delta\left(W_1 \textsf{ChannelStat}(\mathbf{X}_{\textsf{in}})\right)\right)
    \label{eq:affinity}
\end{equation}

where $\delta$ and $\sigma$ represent non-linearity functions, and $W_1$ and $W_2$ are learnable weights. The role of $\sigma$ is to assign an importance weight to each channel to compute the affinity. The result of the affinity calculation is used to recalibrate the output of the Grouped Mamba operator, as shown in Eq.~\ref{eq:recalib}.
\begin{equation}
    \mathbf{X}_{\textsf{CAM}} = \textsf{CAM}(\mathbf{X}_{\textsf{GM}}, \textsf{Affinity}(\mathbf{X}_{\textsf{in}})) = \mathbf{X}_{\textsf{GM}} \cdot \textsf{Affinity}(\mathbf{X}_{\textsf{in}})
    \label{eq:recalib}
\end{equation}
where $\mathbf{X}_{\textsf{CAM}}$ is the recalibrated output, $\mathbf{X}_{\textsf{GM}}$ is the concatenated output of the four VSSS groups from Eq.~\ref{eq:groupedmamba}, $\mathbf{X}_{\textsf{in}}$ is the input tensor, and $\textsf{Affinity}(\mathbf{X}_{\textsf{in}})$ are the channel-wise attention scores obtained from the channel affinity calculation operation in Eq.~\ref{eq:affinity}.

While the average pooling and affinity procedure employed by the CAM module resembles the Squeeze-and-Excitation (SE) block~\cite{hu2018squeeze}, it introduces a distinct mechanism tailored explicitly for cross-channel attention within multi-group transformations. Specifically, CAM allows inter-group information exchange to overcome the inherent limitations of the "Grouped Mamba Operator," which inherently restricts interactions within individual groups. In contrast, SE blocks typically focus on recalibrating a single feature group and have not yet been investigated within the context of Mamba-based architectures.

\subsection{Distilled Loss Function}
\label{sec:method:loss}

As mentioned earlier in the motivation in Sec.~\ref{sec:method}, the Mamba training is unstable when scaled to large models~\citep{patro2024simba}. To mitigate this issue, we propose to utilize a distillation objective alongside the standard cross-entropy objective. Knowledge distillation involves training a student model to learn from a teacher model's behavior by minimizing a combination of the classification loss and distillation loss. The distillation loss is computed using the cross-entropy objective between the logits of the teacher and student models. Given the logits ($Z_s$) from the student model, logits ($Z_t$) from a teacher model (RegNetY-16G~\citep{radosavovic2020regnet} in our case), the ground truth label $y$, and the hard decision of the teacher $y_\mathrm{t}=\mathrm{argmax}_c Z_\mathrm{t}(c)$, the joint loss function is defined as shown in Eq.~\ref{eq:loss}.

\begin{equation}
    \mathcal{L}_\mathrm{total} = \alpha\mathcal{L}_\mathrm{CE}(Z_s,y) + (1-\alpha)\mathcal{L}_\mathrm{CE}(Z_s,y_\mathrm{t}).
    \label{eq:loss}
\end{equation}

where $\mathcal{L}_\mathrm{CE}$ is the cross-entropy objective and $\alpha$ is the weighting parameter. We demonstrate in the supplementary material that incorporating the distilled loss enhances training stability, resulting in consistent performance improvements for larger model variants.
\section{Experiments}
\label{sec:results}

\begin{table*}[t]
\centering
\setlength{\tabcolsep}{0.5cm}
    \begin{tabular}{c|cccc|c}
    \toprule
        Method & \begin{tabular}[c]{@{}c@{}}Token \\ mixing\end{tabular} & \begin{tabular}[c]{@{}c@{}}Image \\ size\end{tabular} & \#Param. & FLOPs &  Top-1 acc. \\
    \midrule
        RegNetY-8G~\citep{radosavovic2020regnet} & Conv & 224$^2$ & 39M & 8.0G  &  81.7 \\
        RegNetY-16G~\citep{radosavovic2020regnet} & Conv  & 224$^2$ &  84M & 16.0G & 82.9 \\
    \midrule
        EffNet-B4~\citep{tan2019efficientnet} & Conv & 380$^2$ & 19M & 4.2G  &  82.9 \\
        EffNet-B5~\citep{tan2019efficientnet} &Conv  & 456$^2$ & 30M & 9.9G &  83.6 \\
    \midrule
        DeiT-S~\citep{touvron2021deit} & Attention & 224$^2$ & 22M & 4.6G  &  79.8 \\
        DeiT-B~\citep{touvron2021deit} & Attention & 224$^2$ & 86M & 17.5G  &  81.8 \\
        DeiT-B~\citep{touvron2021deit} & Attention & 384$^2$ & 86M & 55.4G  &  83.1 \\
    \midrule
        ConvNeXt-T~\citep{liu2022convnext} & Conv & 224$^2$ & 29M & 4.5G &  82.1 \\
        ConvNeXt-S~\citep{liu2022convnext} & Conv & 224$^2$ & 50M & 8.7G &   83.1 \\
        ConvNeXt-B~\citep{liu2022convnext} &Conv &  224$^2$ & 89M & 15.4G &  83.8 \\
    \midrule
        Swin-T~\citep{liu2021Swin} & Attention & 224$^2$ & 28M & 4.6G &  81.3 \\
        Swin-S~\citep{liu2021Swin} & Attention & 224$^2$ & 50M & 8.7G &  83.0 \\
        Swin-B~\citep{liu2021Swin} & Attention & 224$^2$ & 88M & 15.4G &  83.5 \\
    \midrule
        ViM-S~\citep{lianghui2024vim} &  SSM & 224$^2$ & 26M & -  &  80.5 \\
        VMamba-T~\citep{yue2024vmamba} &  SSM & 224$^2$ & 31M & 4.9G  &  82.5 \\
        VMamba-S~\citep{yue2024vmamba} &  SSM & 224$^2$ & 50M & 8.7G &  83.6 \\
        VMamba-B~\citep{yue2024vmamba} &  SSM & 224$^2$ & 89M & 15.4G &  83.9 \\
        LocalVMamba-T~\citep{huang2024localmamba} &  SSM & 224$^2$ & 26M & 5.7G  &  82.7 \\
        LocalVMamba-S~\citep{huang2024localmamba} &  SSM & 224$^2$ & 50M & 11.4G &  83.7 \\
        EfficientVMamba-B~\citep{pei2024efficientvmamba} &  SSM & 224$^2$ & 33M & 4.0G & 81.8 \\
    \midrule
        \rowcolor{front-color}
        \rowcolor{front-color}
        GroupMamba-T&  SSM & 224$^2$ & 23M & 4.5G &  83.3 \\
        \rowcolor{front-color}
        \rowcolor{front-color}
        GroupMamba-S&  SSM & 224$^2$ & 34M & 7.0G & 83.9\\
        \rowcolor{front-color}
        \rowcolor{front-color}
        GroupMamba-B& SSM & 224$^2$ & 57M & 14G  & 84.5 \\

    \bottomrule
    \end{tabular}
    \caption{\textbf{Performance comparison of GroupMamba models with state-of-the-art convolution-based, attention-based, and SSM-based models on ImageNet-1K~\citep{deng2009imagenet}}. Our models demonstrate better trade-off between accuracy and parameters.} 
\label{tab:imagenet}
\end{table*}

\subsection{Image Classification }
\textbf{Settings:} The image classification experiments are based on ImageNet-1K~\citep{deng2009imagenet}, which comprising of over $1.28$ million training images and 50K validation images, spanning $1,000$ categories. Following~\cite{liu2022swinv2}, we train our models for using the AdamW~\citep{loshchilov2018adamw} optimizer and a cosine decay learning rate scheduler for $300$ epochs, including a $20$ epoch warm-up. The total batch size is set to $1024$, with models trained on 8x A100 GPUs, each with 80GB of CUDA memory. Optimizer betas are set to $(0.9, 0.999)$; momentum is set to $0.9$, and an initial learning rate of $1\times10^{-3}$ is used with a weight decay of $0.05$. Label smoothing of $0.1$ is used alongside the distillation objective (see Sec.~\ref{sec:method:loss}).

\begin{table*}[htp!]
\centering
\setlength{\tabcolsep}{0.13cm}
    \begin{tabular}{c|ccc|ccc|cc|cc}
    \toprule
    \multicolumn{9}{c}{\textbf{Detection \& Instance Segmentation}} & \multicolumn{2}{c}{\textbf{Semantic Segmentation}} \\
    \midrule
        Backbone & AP$^\text{b}$ & AP$^\text{b}_\text{50}$ & AP$^\text{b}_\text{75}$ & AP$^\text{m}$ & AP$^\text{m}_\text{50}$ & AP$^\text{m}_\text{75}$ & \#param. & FLOPs & mIoU (SS) & mIoU (MS) \\
    \midrule
        ResNet-50~\citep{he2016resnet} & 38.2 & 58.8 & 41.4 & 34.7 & 55.7 & 37.2 & 44M & 260G & 42.1 & 42.8 \\
        Swin-T~\citep{liu2021Swin} & 42.7 & 65.2 & 46.8 & 39.3 & 62.2 & 42.2 & 48M & 267G & 44.4 & 45.8 \\
        ConvNeXt-T~\citep{liu2022convnext} & 44.2 & 66.6 & 48.3 & 40.1 & 63.3 & 42.8 & 48M & 262G & 46.0 & 46.7 \\
        VMamba-T~\citep{yue2024vmamba} & 47.4 & 69.5 & 52.0 & 42.7 & 66.3 & 46.0 & 50M & 270G & 48.3 & 48.6 \\
        LocalVMamba-T~\citep{huang2024localmamba} & 46.7 & 68.7 & 50.8 & 42.2 & 65.7 & 45.5 & 45M & 291G & 47.9 & 49.1 \\
        \rowcolor{front-color}
        GroupMamba-T & 47.6 & 69.8 & 52.1 & 42.9 & 66.5 & 46.3 & 40M & 279G & 48.6 & 49.2 \\
    
    \bottomrule
\end{tabular}
\caption{\textbf{Comparison of model performance on dense prediction tasks}: Object detection and instance segmentation results on MS-COCO~\citep{lin2017coco} using Mask R-CNN 1$\times$ schedule~\citep{he2017mask}, and semantic segmentation results on ADE20K~\citep{zhou2017ade20k} using UperNet~\citep{xiao2018unified}. 'SS' and 'MS' denote single-scale and multi-scale evaluations, respectively. $AP^{b}$ and $AP^{m}$ represent box and mask AP.}
\label{tab:coco_eval}
\end{table*}

\noindent\textbf{Results:} 
Tab.~\ref{tab:imagenet} presents a comparison of our proposed GroupMamba models (T, S, B) with various state-of-the-art methods. The GroupMamba models exhibit a notable balance of accuracy and computational efficiency.
GroupMamba-T achieves a top-1 accuracy of $83.3\%$ with $23$ million parameters and $4.5$ GFLOPs, outperforming ConvNeXt-T~\citep{liu2022convnext} and Swin-T~\citep{liu2021Swin} by $1.2\%$ and $2.0\%$, respectively, with fewer parameters. Additionally, GroupMamba-T surpasses the recently introduced SSM models, outperforming VMamba-T~\citep{yue2024vmamba} and LocalVMamba-T~\citep{huang2024localmamba} by $0.8\%$ and $0.6\%$, respectively, while using $26\%$ fewer parameters than VMamba-T.
GroupMamba-S, with $34$ million parameters and $7.0$ GFLOPs, achieves an accuracy of $83.9\%$, surpassing VMamba-S~\citep{yue2024vmamba}, Swin-S~\citep{liu2021Swin}, and EfficientVMamba-B~\citep{pei2024efficientvmamba}. The performance is better than LocalVMamba-S~\citep{huang2024localmamba} by 0.2\% with $32\%$ fewer parameters.
Furthermore, GroupMamba-B achieves an accuracy of $84.5\%$ with only $57$ million parameters and $14$ GFLOPs, exceeding VMamba-B~\citep{yue2024vmamba} by $0.6\%$ while using $36\%$ fewer parameters. 


\subsection{Object Detection and Instance Segmentation}

\textbf{Settings:} We evaluate the performance of GroupMamba-T for object detection on the MS-COCO 2017 dataset~\citep{lin2017coco}. Our method is based on the Mask R-CNN 1$\times$ schedule~\citep{he2017mask} detector with the hyperparameters as used for Swin~\citep{liu2021Swin}. We use the AdamW~\citep{loshchilov2018adamw} optimizer and train Mask-RCNN with GroupMamba-T backbone for $12$ epochs. The backbone is initialized and fine-tuned from the ImageNet-1K~\citep{deng2009imagenet}. We use an initial learning rate of $1\times10^{-4}$ and decay by a factor of $10$ at epochs $9$ and $11$. FLOPs are computed for an input dimension of $1280\times800$.

\begin{figure*}
  \centering
    \includegraphics[width=0.97\textwidth]
    {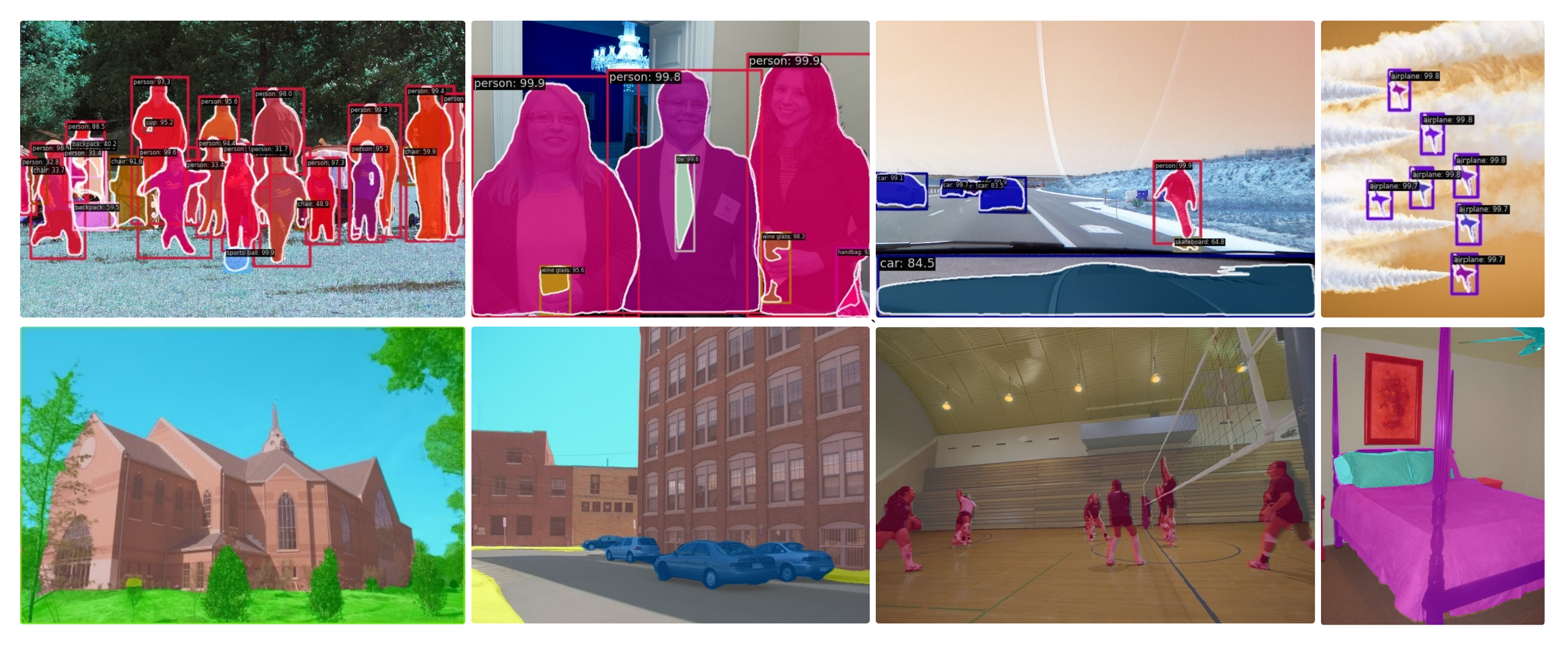}
    \caption{Qualitative results of GroupMamba-T for object detection and instance segmentation (first row) on the MS-COCO val. set and semantic segmentation (second row) on ADE20k val. set.}
    \label{fig:qualitative_fig}
\end{figure*}

\noindent\textbf{Results:}
Tab.~\ref{tab:coco_eval} shows the results of GroupMamba-T, comparing it against various state-of-the-art models for object detection and instance segmentation using the Mask R-CNN framework on the MS-COCO dataset. Our model achieves box AP (AP$^\text{b}$) of 47.6 and mask AP (AP$^\text{m}$) of 42.9. It surpasses ResNet-50~\citep{he2016resnet}, Swin-T~\citep{liu2022swinv2}, ConvNeXt-T~\citep{liu2022convnext}. In addition, GroupMamba-T has competitive performance compared to VMamba-T~\citep{yue2024vmamba} and LocalVMamba-T~\citep{huang2024localmamba}, with less 20\% parameters compared to VMamba-T.
Fig.~\ref{fig:qualitative_fig} (first row) displays qualitative examples of object detection and instance segmentation. GroupMamba-T accurately detects and segments the targets in various scenes. More qualitative examples are presented in the supplementary material.


\subsection{Semantic Segmentation}

\textbf{Settings:} We also evaluate the performance of GroupMamba-T for semantic segmentation on the ADE20K~\citep{zhou2017ade20k} dataset. The framework is based on the UperNet~\citep{xiao2018unified} architecture, and we follow the same hyperparameters as used for the Swin~\citep{liu2021Swin} backbone. More specifically, we use the AdamW~\citep{loshchilov2018adamw} optimizer for a total of $160k$ iterations with an initial learning rate of $6\times10^{-5}$. The default resolution in our experiments is $512\times512$. 

\noindent\textbf{Results:} The GroupMamba-T model demonstrates favorable performance in semantic segmentation compared to various state-of-the-art methods, as presented in Tab.~\ref{tab:coco_eval}. GroupMamba-T achieves a mIoU of $48.6$ in single-scale and $49.2$ in multi-scale evaluation. This outperforms ResNet-50~\citep{he2016resnet}, Swin-T~\citep{liu2021Swin}, and ConvNeXt-T~\citep{liu2022convnext}. Additionally, GroupMamba-T exceeds the performance of the recent SSM methods, including ViM-S~\citep{lianghui2024vim},  VMamba-T~\citep{yue2024vmamba}, and LocalVMamba~\citep{huang2024localmamba}. Fig.~\ref{fig:qualitative_fig} (second row) shows qualitative examples of GroupMamba-T. These examples demonstrate our model's ability to accurately segment various classes for indoor and outdoor scenes. More qualitative examples are presented in the supplementary material.


\begin{figure*}[htp]
  \centering
    \includegraphics[width=1.0\textwidth]
    {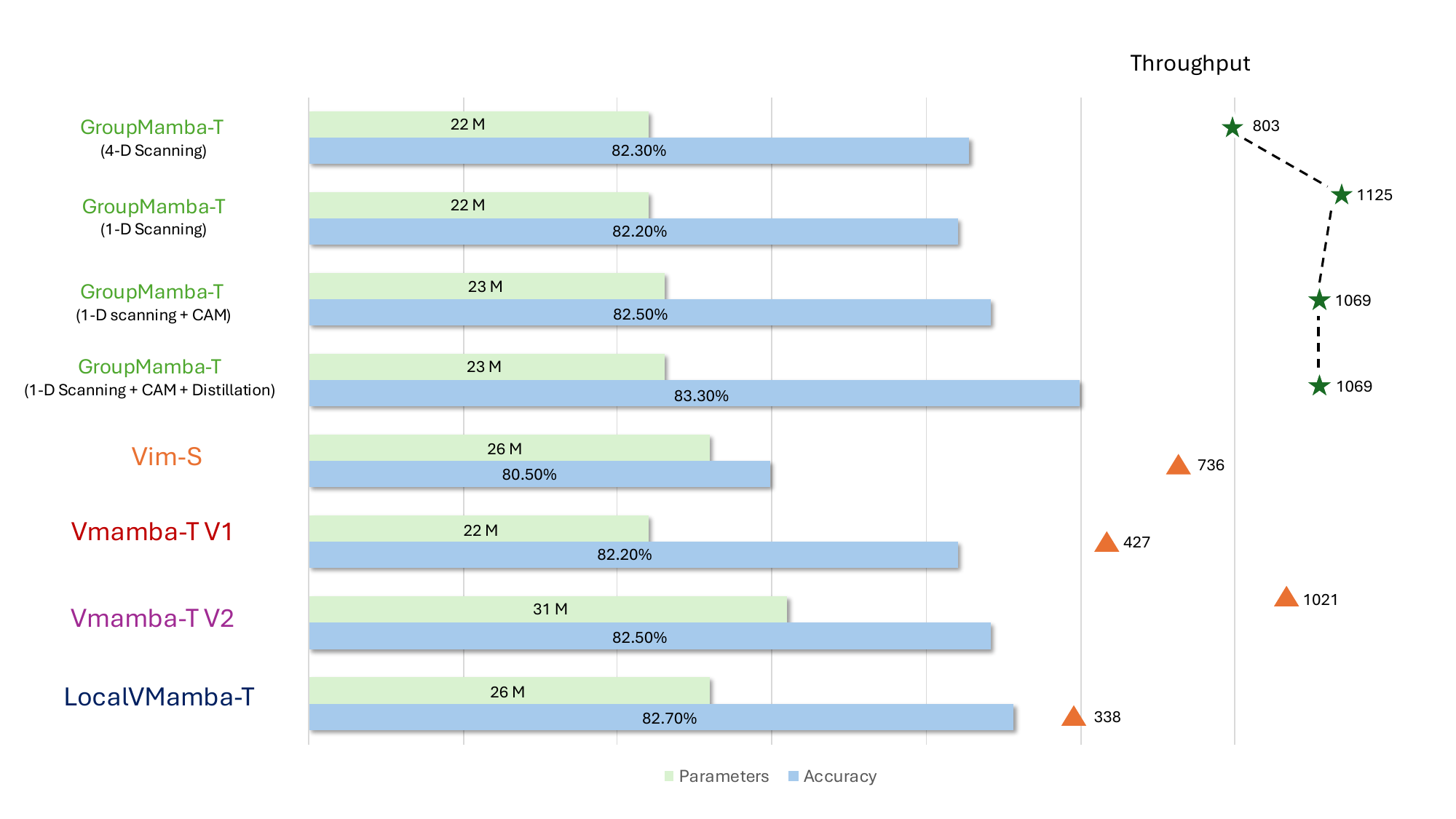}
    \caption{Comparison of GroupMamba variants and SSM-based methods in top-1 accuracy on ImageNet-1k~\cite{deng2009imagenet} and computational efficiency in terms of throughput and number of parameters. The throughput (number of predicted samples per second) is measured using a single NVIDIA A100 GPU with a batch size of 128 for all methods.}
   
    \label{fig:roadmap_figure}
\end{figure*}

\subsection{Ablation Study}

Fig.~\ref{fig:roadmap_figure} shows the impact of each proposed contribution in terms of top-1 accuracy, number of parameters, and throughput, compared to other SSM-based methods. GroupMamba-T with 4-D scanning, comprising 22M parameters, achieves a top-1 accuracy of 82.30\% and a throughput of 803. By applying a unidirectional 1D scan across  $N/4$ channels in four directions—left-to-right, right-to-left, top-to-bottom, and bottom-to-top instead of the full 4-D scanning across all $N$ channels, the throughput significantly increased from 803 to 1125, with only 0.1\% drop in accuracy and the same number of parameters.

The integration of the CAM module further elevates the top-1 accuracy from 82.20\% to 82.50\%, with a minor reduction in throughput (from 1125 to 1069). Finally, incorporating the proposed distillation-based loss pushes the top-1 accuracy to 83.30\%, while preserving the throughput at 1069. Compared to Vim-S~\citep{lianghui2024vim}, GroupMamba-T demonstrates a more efficient design, achieving a 2.8\% improvement in top-1 accuracy with 1.5$\times$ higher throughput, all while utilizing fewer parameters. Compared to LocalVMamba-T~\citep{huang2024localmamba}, GroupMamba-T has a 0.6\% higher accuracy in top-1 accuracy, with 3$\times$ faster and smaller number of parameters. Regarding VMamba-T V1~\citep{yue2024vmamba}, our model achieves a 1.1\% gain in top-1 accuracy with a comparable number of parameters while being faster by 2.5$\times$. Likewise, when compared to VMamba-T V2~\citep{yue2024vmamba}, GroupMamba-T shows marginally faster throughput, an increase of 0.8\% in top-1 accuracy, and a 26\% improvement in parameter efficiency. 
\section{Conclusion and Future Work}
\label{sec:conclusion}

In this paper, we tackle the computational inefficiencies and stability challenges associated with visual SSMs for computer vision tasks by introducing a novel layer called \emph{Modulated Group Mamba}. We also propose a multi-directional scanning method that improves parameter efficiency by scanning in four spatial directions and leveraging \emph{Channel Affinity Modulation} (CAM) operator to enhance feature aggregation across channels. To stabilize training, especially for larger models, we employ a distillation-based training objective. Our experiments demonstrate that the proposed GroupMamba models outperform recent SSMs while being more efficient in terms of parameters and throughput.

Our research has focused on image classification, object detection, and segmentation. To further validate and extend the generalization ability of our method, we aim to explore additional downstream tasks, such as video recognition and time-series data applications. Evaluating the Modulated Group Mamba layer in these contexts will help to uncover its potential benefits and limitations, providing deeper insights and guiding further improvements.

\section{Acknowledgments}
The computations were enabled by resources provided by NAISS at Alvis
partially funded by Swedish Research Council through grant agreement no. 2022-06725, LUMI
hosted by CSC (Finland) and LUMI consortium, and by Berzelius resource provided by the Knut and
Alice Wallenberg Foundation at the NSC.

Syed Talal Wasim and Juergen Gall have been supported by the Federal Ministry of Education and Research (BMBF) under grant no. 01IS22094A WEST-AI and the ERC Consolidator Grant FORHUE (101044724).

\bibliographystyle{ieeenat_fullname}
\bibliography{main}
\clearpage
\begin{center}
\textbf{{\Large Supplementary Material}}
\end{center}

\noindent
In this section, we further include more results and analysis to complement the main paper. We provide additional details on the following topics:
\begin{itemize}
    \item Architectural Details (Sec.~\ref{sup:arch_details})
    \item Ablations (Sec.~\ref{sup:ablations})
    \item Qualitative Results (Sec.~\ref{sup:qualtitative_results})
    \item Discussion (Sec.~\ref{sup:discussion})
    \item Limitations (Sec.~\ref{sup:limitation})
\end{itemize}

\section{Architectural Details}
\label{sup:arch_details}
We develop three variants of our GroupMamba backbones, each tailored to different performance and efficiency requirements: GroupMamba-T (Tiny), GroupMamba-S (Small), and GroupMamba-B (Base), with 23M, 34M, and 57M parameters, respectively. These variants differ in their channel dimensions and the number of layers per stage, as detailed in Tab.~\ref{tab:archs}.

\section{Ablations}
\label{sup:ablations} 

In Tab.~\ref{tab:ablation_2}, we provide additional ablation results regarding the distillation training objective. For the GroupMamba-T and GroupMamba-S variants, the distilled loss improves performance by an absolute gain of 0.8\% and 0.9\%, respectively. For the largest variant, GroupMamba-B, the distilled loss improves performance by 1.3\%. This demonstrates that larger Mamba-based models with MLP tend to saturate and struggle to converge effectively without distillation. Incorporating distillation for the large model boosts its performance from 83.2\% to 84.5\%.

\begin{figure}[t]
  \begin{center}
    \includegraphics[width=0.47\textwidth]
    {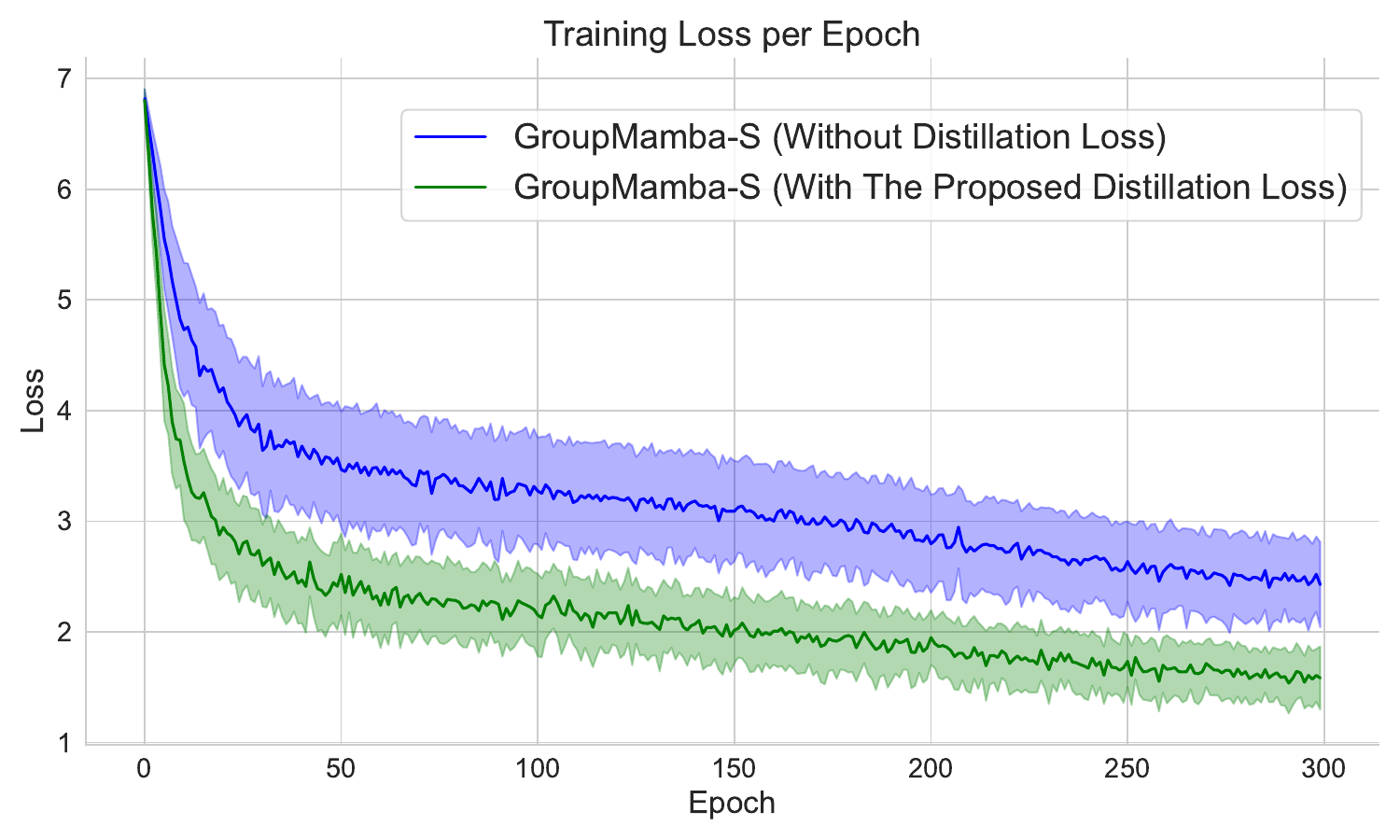}
  \end{center}
  \caption{Training loss visualization for GroupMamba-S with and without the proposed distilled loss.}
  \label{fig:loss_visualization}
\end{figure}

We also visualize the training loss curves with and without our proposed distilled loss for GroupMamba-S in Fig.~\ref{fig:loss_visualization}. The shaded areas indicate the standard deviation of loss across the training epochs. As shown, incorporating the distilled loss (green curve) consistently leads to lower training losses and less loss variability throughout the training process, leading to improved stability.

We compare in Tab.~\ref{tab:abl_2} the performance of different scanning directions with respect to the number of groups for GroupMamba-T. In the first row, we use Direction 1. In the second row, we use Direction 1 and Direction 2. In the last row, we use the four scanning scanning directions (As visualized in Fig.~\ref{fig:method_overall} (d). Four groups with four directions capture richer spatial cues, which provide comprehensive feature representation and lead to higher top-1 accuracy with comparable throughput. 

We also conduct an ablation study to evaluate efficiency with varying numbers of groups. While utilizing two groups reduces parameters by 15\% and four groups achieves a reduction of 26\%, employing eight groups yields only a marginally greater reduction of 28\% due to the nonlinear scaling of MLP parameters. In addition, using eight groups (with eight scanning directions) decreases throughput, negatively impacting model efficiency. Hence, four groups have the optimal trade-off between parameter reduction and high throughput.

\begin{table}[t]
\centering
\resizebox{\columnwidth}{!}{%
\begin{tabular}{c|ccc}
    \toprule
    Method & \#Param. & FLOPs &  Top-1 acc.  \\
    \midrule
     GroupMamba-T w/o Distilled Loss  & 23M & 4.6G  &  82.5 \\
     GroupMamba-T with Distilled Loss & 23M & 4.6G & 83.3 \scriptsize{(+0.8)}\\
     \midrule
    GroupMamba-S w/o Distilled Loss  & 34M &  7.0G   &  83.0 \\
     GroupMamba-S with Distilled Loss & 34M &  7.0G  & 83.9 \scriptsize{(+0.9)}\\
     \midrule
     GroupMamba-B w/o Distilled Loss  & 57M & 14G  &  83.2 \\
     GroupMamba-B with Distilled Loss & 57M & 14G & 84.5 \scriptsize{(+1.3)}\\
    \bottomrule
\end{tabular}
}
\caption{Ablation study on GroupMamba variants with and without the Distilled Loss.}
\label{tab:ablation_2}
\end{table}
In Tab.~\ref{tab:comparison}, we present an additional ablation study and a fair comparison between GroupMamba-T and VMamba-T without distillation alongside another variant of GroupMamba-T designed to match the parameter count of VMamba-T for balanced evaluation. Remarkably, GroupMamba-T achieves equivalent performance to VMamba-T with 26\% fewer parameters. When parameter counts are matched, the enhanced variant, GroupMamba-T\textsuperscript{\textdagger}, outperforms VMamba-T, achieving a top-1 accuracy of 83.1\% on ImageNet-1K, compared to 82.5\% for VMamba-T, without using any distillation.

\begin{table}[t]
  \centering
  \resizebox{\columnwidth}{!}{%
  \begin{tabular}{lccc}
    \toprule
    Scanning Directions & Throughput (ms) $\uparrow$ & \#Param $\downarrow$ & Top-1 (\%) $\uparrow$ \\ 
    \midrule
     D1                    & 1096 & 23M & 82.9 \\
    D1, D2                & 1087 & 23M & 83.1 \\
    \textbf{D1, D2, D3, D4} & \textbf{1069} & \textbf{23M} & \textbf{83.3} \\ 
    \bottomrule
  \end{tabular}
  }
  \caption{Comparison of different scanning directions in terms of throughput, parameters, and top-1 accuracy for GroupMamba-T.}
  \label{tab:abl_2}
\end{table}

\section{Qualitative Results}
\label{sup:qualtitative_results}
\begin{figure}
  \centering
    \includegraphics[width=0.5\textwidth]
    {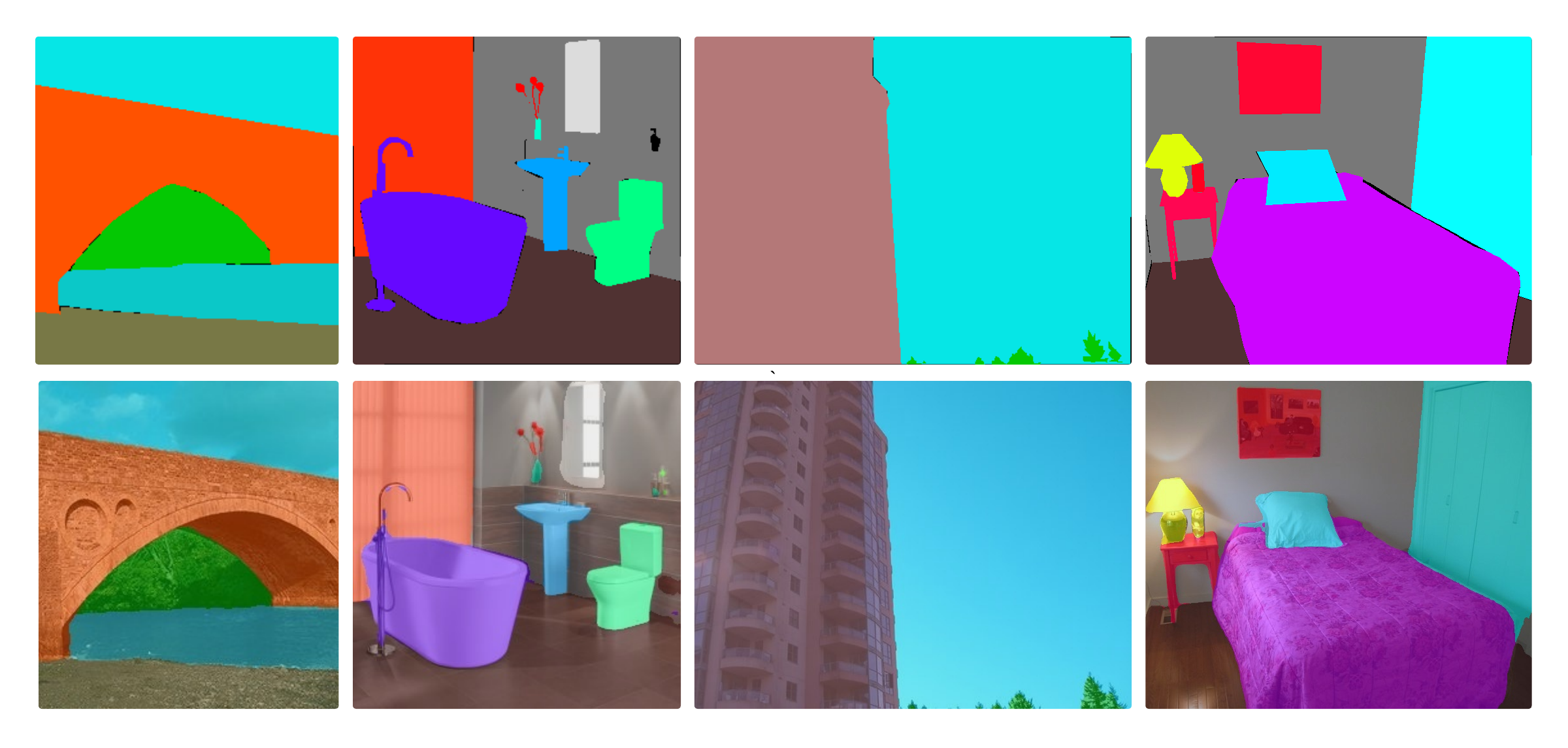}
    \caption{Qualitative results of GroupMamba-T for semantic segmentation on ADE20K validation set. The first row shows the ground truth for the masks, while the second and second show the corresponding predictions of our model.}
    \label{fig:segmentation_fig}
\end{figure}

\begin{figure*}
  \centering
    \includegraphics[width=0.76\textwidth]
    {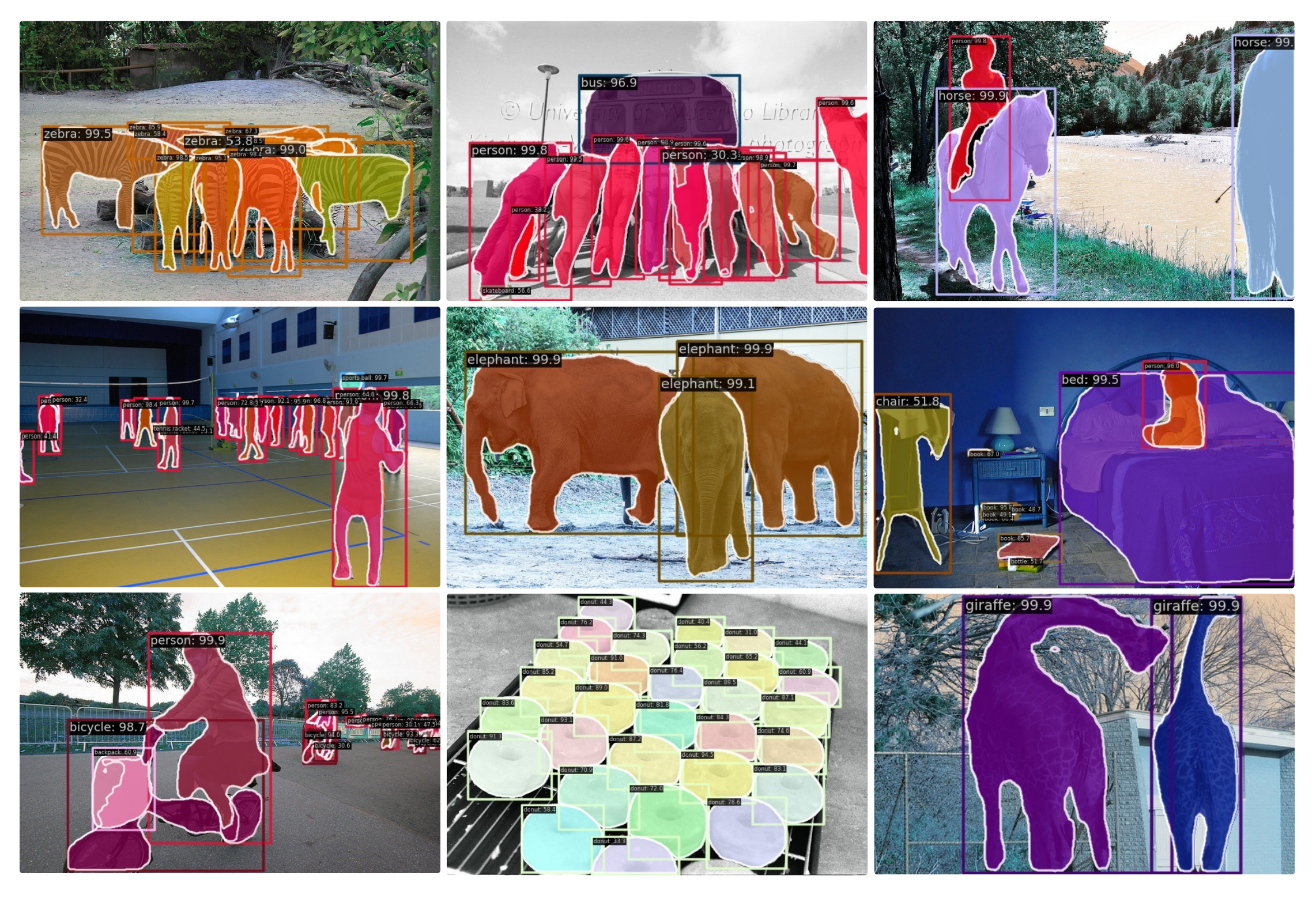}
    \caption{Qualitative results of GroupMamba-T for object detection and instance segmentation on the COCO validation set.}
    \label{fig:detection_fig}
\end{figure*}

In Fig.~\ref{fig:segmentation_fig}, we show additional qualitative results of GroupMamba-T on samples from the ADE20K~\cite{zhou2017ade20k} validation set for semantic segmentation. The first row shows the ground truth masks, while the second row displays the predicted masks. Our model consistently has sharp and accurate delineations, effectively capturing fine details and complex object boundaries, further emphasizing its robustness in semantic segmentation. Similarly, we present in Fig.~\ref{fig:detection_fig} additional qualitative results of GroupMamba-T on samples from the COCO validation set~\cite{lin2017coco}, showcasing its strong performance in both instance segmentation and object detection tasks. The model excels at accurately localizing objects and producing precise segmentations, even in complex scenes with varying scales, multiple instances, and challenging backgrounds. The quantitative and qualitative results of GroupMamba demonstrate the robust generalization capability of our GroupMamba backbones across diverse downstream tasks, including semantic segmentation, object detection, and instance segmentation.

\begin{table}[t!]
\centering
\setlength{\tabcolsep}{0.2cm}
\begin{tabular}{c|ccc}
    \toprule
    Method & \#Param. & FLOPs &  Top-1 acc.  \\
    \midrule
    VMamba-T  & 31M & 4.9G  & 82.50\% \\
    GroupMamba-T& 23M & 4.5G & 82.50\%\\
    GroupMamba-T\textsuperscript{\textdagger}& 31M & 5.2G & 83.10\%\\
    \bottomrule
\end{tabular}
\caption{Comparison of VMamba-T and GroupMamba-T without distillation. The number of channels is increased in GroupMamba-T\textsuperscript{\textdagger} to match the same parameters of VMamba-T}
\label{tab:comparison}
\end{table}

\begin{table*}[t]
\centering
\small
\setlength{\tabcolsep}{7.5pt}
\renewcommand\arraystretch{1.0}

\begin{tabular}{c|c|c|c|c|c|c}
\toprule
\multirow{2}{*}{Stage} & \multirow{2}{*}{Output Resolution} & \multirow{2}{*}{Type} & \multirow{2}{*}{Config} & \multicolumn{3}{c}{GroupMamba}  \\ \cline{5-7}
                       &                                    &                       &                         & T         & S        & B       \\
\hline
\hline
\multirow{4}{*}{stem}  & \multirow{2}{*}{$\frac{H}{2}\times \frac{W}{2}$} & \multirow{2}{*}{\begin{tabular}[c]{@{}c@{}}Patch\\ Embed.\end{tabular}} & Patch Size & \multicolumn{3}{c}{$k=3\times 3, s=2$}      \\ \cline{4-7}
                       &                                    &                       & Embed. Dim.             & 32         & 64        & 64      \\ 
                       \cline{2-7}
                       & \multirow{2}{*}{$\frac{H}{4}\times \frac{W}{4}$} & \multirow{2}{*}{\begin{tabular}[c]{@{}c@{}}Patch\\ Embed.\end{tabular}} & Patch Size & \multicolumn{3}{c}{$k=3\times 3, s=2$}      \\ \cline{4-7}
                       &                                    &                       & Embed. Dim.             & 96         & 96        & 128      \\
                       \hline 
\multirow{1}{*}{1}     & \multirow{1}{*}{$\frac{H}{4}\times \frac{W}{4}$} & \multirow{1}{*}{$\texttt{Modulated Group Mamba}$} & Stage [$C_1$, $N_1$] & 96, 2 & 96, 2 & 128, 2  \\ 
                       \hline 
\multirow{5}{*}{2}     & \multirow{2}{*}{$\frac{H}{8}\times \frac{W}{8}$} & \multirow{2}{*}{\begin{tabular}[c]{@{}c@{}}Down-sampling\end{tabular}} & Patch Size  & \multicolumn{3}{c}{$k=3\times 3, s=2$}      \\ \cline{4-7}
                       &                                    &                       & Embed. Dim.             & 192         & 192       & 256      \\
                       \cline{1-7}
                       &   \multirow{1}{*}{$\frac{H}{8}\times \frac{W}{8}$} & \multirow{1}{*}{$\texttt{Modulated Group Mamba}$} & Stage [$C_2$, $N_2$] & 192, 2 & 192, 2 & 256, 2  \\ 
                       \hline 
\multirow{5}{*}{3}     & \multirow{2}{*}{$\frac{H}{16}\times \frac{W}{16}$} & \multirow{2}{*}{\begin{tabular}[c]{@{}c@{}}Down-sampling\end{tabular}} & Patch Size  & \multicolumn{3}{c}{$k=3\times 3, s=2$}      \\ \cline{4-7}
                       &                                    &                       & Embed. Dim.             & 368        & 384      & 496     \\
                       \cline{1-7}
                       &   \multirow{1}{*}{$\frac{H}{16}\times \frac{W}{16}$} & \multirow{1}{*}{$\texttt{Modulated Group Mamba}$} & Stage [$C_3$, $N_3$] & 368, 9 & 384, 20 & 496, 20  \\ 
                       \hline 
\multirow{5}{*}{4}     & \multirow{2}{*}{$\frac{H}{32}\times \frac{W}{32}$} & \multirow{2}{*}{\begin{tabular}[c]{@{}c@{}}Down-sampling\end{tabular}} & Patch Size  & \multicolumn{3}{c}{$k=3\times 3, s=2$}      \\ \cline{4-7}
                       &                                    &                       & Embed. Dim.             & 760        & 768      & 1012     \\
                       \cline{1-7}
                       & \multirow{1}{*}{$\frac{H}{32}\times \frac{W}{32}$} & \multirow{1}{*}{$\texttt{Modulated Group Mamba}$} & Stage [$C_4$, $N_4$] & 760, 2 & 768, 2 & 1012, 2  \\ 

\bottomrule
Parameters  &  &   &  & 23M & 34M & 57M \\
GFLOPs  &  &   &  & 4.5G & 7.0G & 14.0G \\
\bottomrule
\end{tabular}
\caption{\textbf{GroupMamba Architectures.} Description of the configurations of each model with respect to the output resolution, the output channels $C$, the number of blocks $N$, and the model's GFLOPs and parameters. Between two consecutive stages, a down-sampling layer is used to increase the features and reduce the resolution by two.}
\label{tab:archs}
\end{table*}

\section{Discussion}
\label{sup:discussion}

Our main contributions include introducing the Modulated Group Mamba layer, which enhances computational efficiency and interaction in state-space models through a multi-direction scanning method. We also introduce the Channel Affinity Modulation (CAM) operator to improve feature aggregation across channels, addressing limitations in grouping operations. Additionally, we employ a distillation-based training objective to stabilize the training of models with a large number of parameters. These contributions enable us to achieve competitive performance with recent state-space models in image classification, object detection, instance segmentation, and semantic segmentation with fewer number of parameters.

This can further facilitate the development of vision foundation models based on Mamba that can be scaled to a large number of parameters efficiently and stably. The Modulated Group Mamba layer and CAM operator enhance computational efficiency and feature interaction, allowing models to manage more extensive and complex datasets without excessive resource demands. The distillation-based training objective ensures stability during training, which is crucial for maintaining performance as model sizes increase. Together, these advancements enable the creation of scalable, reliable vision models that can be deployed effectively in various real-world applications.


\section{Limitations}
\label{sup:limitation}

Despite demonstrating clear improvements in efficiency, stability, and accuracy for image classification tasks and fewer parameters for dense prediction tasks, our proposed Modulated Group Mamba layer shows relatively comparable performance on downstream tasks such as object detection and segmentation to VMamba. This minor improvement can be attributed to the more complex nature and diverse requirements of these dense prediction tasks, where the accuracy relies heavily not only on effective global dependency capture but also on more localized spatial feature aggregation and specialized detection or segmentation heads. The proposed model architecture enhances parameter efficiency and global feature modeling through SSM mechanisms, but addressing the intricacies inherent to localization-sensitive tasks may require additional targeted modules or task-specific optimizations.

Although the incorporation of knowledge distillation has successfully improved training stability and yielded performance gain for large-scale models, investigating more efficient or self-guided stabilization approaches would help enhance the model training practicality without requiring auxiliary external teacher models.

\end{document}